\def\year{2022}\relax
\documentclass[letterpaper]{article} 
\usepackage{aaai22}  
\usepackage{times}  
\usepackage{helvet}  
\usepackage{courier}  
\usepackage[hyphens]{url}  
\usepackage{graphicx} 
\usepackage{natbib}  
\usepackage{caption} 
\DeclareCaptionStyle{ruled}{labelfont=normalfont,labelsep=colon,strut=off} 
\newcommand{\trueTitle}{
Mixed-Reality Robot Behavior Replay: A System Implementation
}

\copyrighttext{Presented at the AI-HRI Symposium at AAAI Fall Symposium Series (FSS) 2022}

\usepackage[textsize=tiny]{todonotes}
\usepackage{enumitem}

\title{\trueTitle}

\author{
    Zhao Han,\textsuperscript{\rm 1}\thanks{Most of this work was completed while Zhao Han was affiliated with the University of Massachusetts Lowell.}
    Tom Williams,\textsuperscript{\rm 1}
    Holly A. Yanco\textsuperscript{\rm 2}
}
\affiliations{
    \textsuperscript{\rm 1}MIRRORLab, Department of Computer Science, Colorado School of Mines, 1500 Illinois St., Golden, CO, USA 80401\\
    \textsuperscript{\rm 2}HRI Lab, Department of Computer Science, University of Massachusetts Lowell, 1 University Ave., Lowell, MA, USA 01854\\
    zhaohan@mines.edu, twilliams@mines.edu, holly@cs.uml.edu,
}

\begin{document}

\maketitle

\begin{abstract}
As robots become increasingly complex, they must explain their behaviors to gain trust and acceptance. However, it may be difficult through verbal explanation alone to fully convey information about past behavior, especially regarding objects no longer present due to robots' or humans' actions. Humans often try to physically mimic past movements to accompany verbal explanations. Inspired by this human-human interaction, we describe the technical implementation of a system for past behavior replay for robots in this tool paper. Specifically, we used Behavior Trees to encode and separate robot behaviors, and schemaless MongoDB to structurally store and query the underlying sensor data and joint control messages for future replay. Our approach generalizes to different types of replays, including both manipulation and navigation replay, and visual (i.e., augmented reality (AR)) and auditory replay. Additionally, we briefly summarize a user study to further provide empirical evidence of its effectiveness and efficiency. Sample code and instructions are available on GitHub at \url{https://github.com/umhan35/robot-behavior-replay}.
\end{abstract}

\section{Introduction}

Robots used in domains like collaborative manufacturing, warehousing, and assistive living stand to have benefits such as improving productivity, reducing work-related injuries, and increasing the standard of living.
Yet the increasingly complexity of the manipulation and navigation tasks needed in these domains can be difficult for users to understand, especially when users need to ascertain the reasons behind robot failures.
As such, there is a surge of interest in improving robot understandability by enabling them to explain themselves, e.g., through function annotation \cite{hayes2017improving}, encoder-decoder deep learning framework \cite{amir2018agent}, interpretable task representation \cite{han2021building}, and software architecture \cite{stange2022self}. Different dimensions of robot explanations have also been explored, such as proactive explanations \cite{zhu2020effects}, preferred explanations \cite{han2021need}, and undesired behaviors \cite{stange2020effects}. However, these works focused on explaining a robot's current behaviors.

\begin{figure}[h]
\centering
\includegraphics[width=0.4\columnwidth]{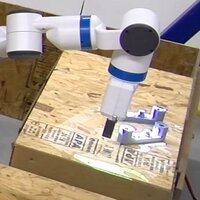}
\caption{Manipulation replay using the replay technique described in this paper. The robot's arm movement and the green projection (bottom) to indicate the object to be grasped were being replayed to clarify a perception failure: A torn-up wood chip was unknowingly misrecognized as one of the gearbox bottoms. Key frames from the same replay and two other types of replays are illustrated in Figure \ref{fig:pick}--\ref{fig:place}.}
\label{fig:firstpage}
\end{figure}

\begin{figure*}[h]
\centering
\newlength{\hpickreplayfig} \setlength{\hpickreplayfig}{1.4in}
\includegraphics[height=\hpickreplayfig]{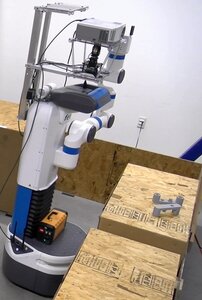}
\includegraphics[height=\hpickreplayfig]{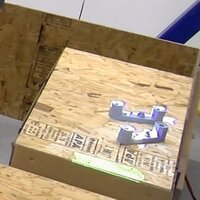}
\includegraphics[height=\hpickreplayfig]{fig-pick-replay/rpp30-above.jpg}
\includegraphics[height=\hpickreplayfig]{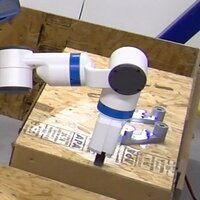}
\includegraphics[height=\hpickreplayfig]{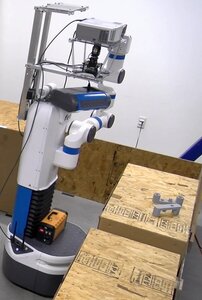}
\caption{\textbf{Manipulation replay of picking a misrecognized object:} 
Start, perceive, reach above, pick, reset. Both arm movement and AR visualizations are replayed. The rectangular green area (bottom) shows the grasped object. White area, projected onto the two gearbox bottoms, shows correctly recognized objects. (\textbf{Video}: \url{https://youtu.be/pj7-LqEsb94})}
\label{fig:pick}
\end{figure*}

\begin{figure*}
\centering
\newlength{\hnavreplayfig} \setlength{\hnavreplayfig}{1.45in}
\includegraphics[height=\hnavreplayfig]{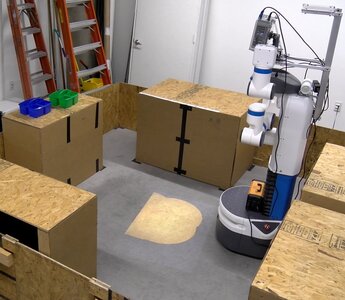}
\includegraphics[height=\hnavreplayfig]{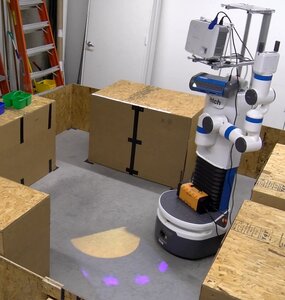}
\includegraphics[height=\hnavreplayfig]{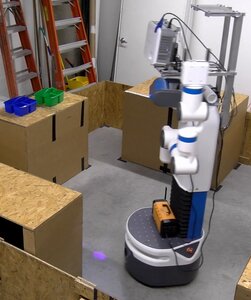}
\includegraphics[height=\hnavreplayfig]{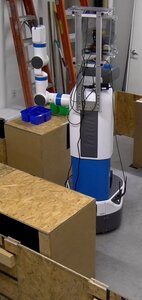}
\includegraphics[height=\hnavreplayfig]{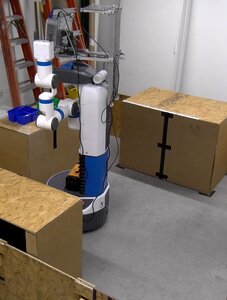}
\caption{\textbf{Navigation replay of a detour path}: Start, rotate, detour, reach position, reach orientation. Both wheel movement and AR visualizations were replayed. Yellow area (spheres of laser scan points; bottom middle) were projected to show ground obstacle, and purple arrows (path poses; bottom) are projected to show past detour path. (\textbf{Video}: \url{https://youtu.be/hV6jsA42YYY})}
\label{fig:nav}
\end{figure*}

\begin{figure*}
\centering
\newlength{\hplacereplayfig} \setlength{\hplacereplayfig}{1.53in}
\includegraphics[height=\hplacereplayfig]{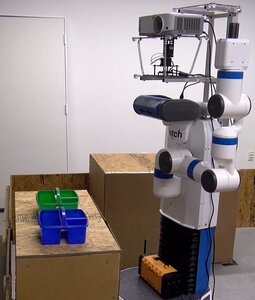}
\includegraphics[height=\hplacereplayfig]{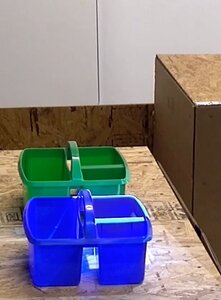}
\includegraphics[height=\hplacereplayfig]{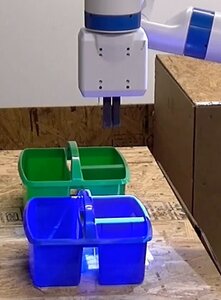}
\includegraphics[height=\hplacereplayfig]{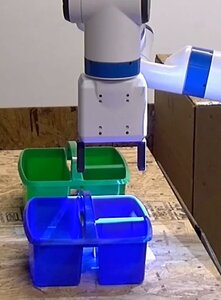}
\includegraphics[height=\hplacereplayfig]{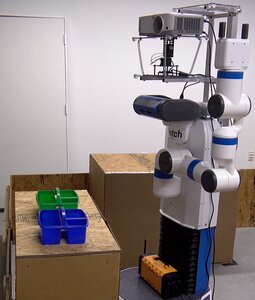}
\caption{\textbf{Manipulation replay of placing (a gearbox bottom) into a wrong caddy section} (the left large section is the correct one): Start, perceive, reach above, place, release, reset. Both arm movement and an AR visualization were replayed. A white cube was projected into the blue caddy's top-right section to show the misrecognized. (\textbf{Video}: \url{https://youtu.be/kIJjU2FR4XU
})}
\label{fig:place}
\end{figure*}

One challenge within this space is enabling robots to explain their past behavior after their environment has changed. This is an interesting yet challenging problem because objects present in the past might have already been replaced or removed from the scene, making the task of referring to those objects during explanation particularly challenging (see also~\citealt{han2022evaluating}). Moreover, a robot may not be capable of reasoning and explaining its past behaviors due to unawareness of failures (see Figure \ref{fig:pick} and \ref{fig:place}), and limited semantic reasoning about objects like ground obstacles or tabletop objects (see also Figure \ref{fig:nav}). 

To help explain a robot's past behaviors, we describe in this tool paper the implementation of a mixed-reality robot behavior replay system that builds on previous work on \textit{Visualization Robots} Virtual Design Elements (VDEs)~\cite{walker2022virtual}. While previous VDEs in this category have primarily sought to visualize future robot behaviors~\cite{rosen2019communicating}, we instead use this technique to visualize previously executed behaviors. %
The robot behaviors that our technique is capable of replaying generalize to replay of both manipulation and navigation behaviors. 
(See Figure \ref{fig:pick}--\ref{fig:place}). Our replay technique can also handle replay of non-physical cues: verbalization, e.g., sound and speech and visualization, such as projector-based augmented reality \cite{han2020projection,han2022projecting}. Empirical evidence of the effectiveness and efficiency of our approach in explaining past behavior has been presented in our previous work  
\cite{han2022past}. While beyond the scope of this tool paper, 
we will briefly mention the experimental results in Section \ref{sec:eval}.

We demonstrate our technique on a mobile manipulator Fetch robot \cite{wise2016fetch} using the widely-used Robot Operating System (ROS) \cite{quigley2009ros}, with the robot behavior encoded in hierarchical behavior trees \cite{colledanchise2018behavior}. Our use of ROS means that our implementation is more-or-less platform agnostic, as most current robots used in research and development have ROS support \cite{rosshowcase} or bridges \cite{scheutz2019overview}.

This work is beneficial to both manipulation and navigation researchers. 
In addition, our replay technique is helpful for visual debugging for robot developers \cite{ikeda2022advancing}, and for explaining past behaviors to non-expert users.

\section{Related Work:\\ Choosing Underlying Technologies}

\subsection{Robot Data Storage}

To replay robot behavior, the first step is to store robot data. One popular tool is rosbag\footnote{\url{https://wiki.ros.org/rosbag}}, which uses filesystems (bag files) to store and play ROS messages. Despite being persistent on disks, 
relying on filesystems, compared to databases that we will discuss soon, made it challenging to query specific behaviors for replaying purposes, because related data in different bag files are unstructured and unlinked, requiring writing custom code and logic.

Thus, roboticists have been exploring database technologies. The schemaless MongoDB database is a popular and justified choice among many researches, e.g., 
\citet*{beetz2010cram,niemueller2012generic, beetz2015open}, to store data from sensors or communication messages. Being schemaless allows storing different data types without creating different data structures for different data messages, such as tables in relational Structured Query Language (SQL) databases, e.g., MySQL. In addition to the large number of different robotics data messages, they are often hierarchical/nested and commonly seen in ROS messages, such as the \textit{PoseStamped} message in the geometry\_msgs package\footnote{\url{https:
//wiki.ros.org/geometry_msgs}}. 
The hierarchical \textit{PoseStamped} message contains a \textit{Header} message to include a reference coordinate frame and a timestamp, and a \textit{Pose} message to include a hierarchical \textit{Point} message for position information and a \textit{Quaternion} message for orientation information. It is imaginably tedious to create all these tables for nested data messages one by one. The advantage of schemaless database is also known as minimal configuration, allowing evolving data structures to support innovation and development \cite{niemueller2012generic}. In this work, we used the \textit{mongodb\_log} library, open-sourced by \citet{niemueller2012generic}, with slight modifications to synchronize timing of different timestamped ROS messages for replay.

\subsection{Robot Behavior Representation}

With a justified choice to use the schemaless MongoDB for robotic data, we then decide how to represent robot behaviors and store related data tied to specific behaviors. A number of methods have been used in prior work to represent sequences of robot actions~\cite{nakawala2018approaches}, including ontologies, state machines, Petri Nets, and behavior trees.

Ontology belongs to the knowledge representation family. Ontologies can be used to infer task specifications from high-level, abstract, underspecified input in a predefined set of actions. Popular implementation include KnowRob \cite{tenorth2009knowrob, beetz2018know} and CRAM \cite{beetz2010cram}. Yet, this approach typically focuses on specification of high-level tasks rather than low-level motion primitives. 

Finite state machines (FSM) are a well-established method for modeling computation \cite{schneider1990implementing} and have been used for robot task specification and execution, such as the SMACH (State MACHine) library \cite{bohren2010smach} and hierarchical RAFCON \cite{brunner2016rafcon}. Although FSM is very flexible at describing task workflow and well-validated, a workflow in FSM can have a significant number of states with intertwined dependencies through transitions, making it hard to maintain, scale and reuse \cite{colledanchise2018behavior}. Particularly for robot behavior replay, it is challenging to clearly separate different robot behaviors.

Petri Nets were created to model concurrency and distributed execution and coordination, which can be seen in multi-robot systems \cite{ziparo2008petri} and soccer robot \cite{palamara2008robotic}. Although useful, we are more interested in sequential actions and thus leave parallel behavior replay to future work.

Finally, behavior trees (BTs) use tree structures to encapsulate behaviors in different kind of parent control nodes with child execution nodes \cite{colledanchise2018behavior}. BTs are commonly used to model AI agents in games \cite{lim2010evolving,sagredo2017trained} and recently have been gaining momentum in robotics, e.g., end-user programming \cite{paxton2017costar}, industrial robots \cite{intera}, learning from demonstration \cite{french2019learning}, and navigation \cite{macenski2020marathon}. Compared to the aforementioned three robot action sequence methods, we choose BTs because they are particularly well suited to represent atomic and separable behaviors as subtrees with control nodes for replay. For a gentle introduction to behavior trees and in a mobile manipulation task, please see our prior work \cite{han2021building}.

\section{Behavior Replay Implementation in ROS}

\subsection{High-Level Workflow}

\begin{figure}
\centering
\includegraphics[width=0.9\columnwidth]{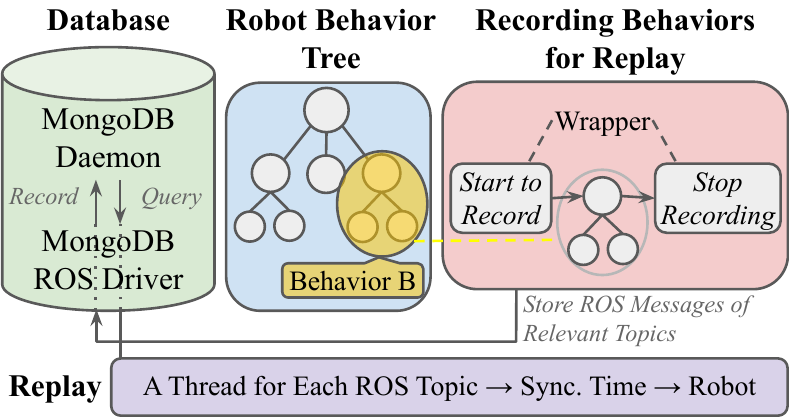}
\caption{A high-level diagram for our robot behavior replay implementation. Robot behaviors are encoded in behavior trees. After Behavior B is identified for future replay, a wrapper is used to record relevant ROS topic to MongoDB through a MongoDB ROS driver. After querying a specific behavior, multiple threads are created for all relevant ROS topics, replaying them after replay time is synchronized.}
\label{fig:workflow}
\end{figure}

As this is a tool paper, now we go through each step in the replay implementation in ROS. Figure \ref{fig:workflow} illustrates them.

First, MongoDB and its ROS driver (See \url{http://wiki.ros.org/mongodb_store}) need to be installed and running.

Second, the robot behaviors should be encoded in behavior trees. Specifically, we have used the \texttt{BehaviorTree.CPP} framework\footnote{\url{https://www.behaviortree.dev/}}. Our sample code shows how behavior nodes are registered and then specified in an XML behavior tree file. In theory, any task specification should work because only the underlying ROS topics are stored in MongoDB database. However, we recommend specifying robot tasks in a behavior tree to have atomic and separable behaviors. With the behavior tree representation, it removes the burden to separate specific behaviors for replay and it is easy to wrap the control node, representing a behavior, to record concerned ROS topics for replay.

Third, the specific behavior represented by the control parent node should be identified for replay. 

Fourth, before and after a control parent node is executed, code for starting and stopping recording, i.e., storing related topics, should be added. This is also called wrapper code and is done through the \texttt{MongodbLogger} class: After initializing an \texttt{MongodbLogger} instance, specific topics for future replay can be set. A number of them will be discussed in the next subsection.

Finally, to query the recorded topics from MongoDB database, one could simply replay the whole collection (analogous to a table in SQL databases) or query by time or topics. Under the hood, a thread is first created for the replay of every ROS topic after they are retrieved, and after a common clock is established, different ROS messages are replayed at the right timestamp to replicate the movements or visualizations that happened at record time. The code for querying and replaying is in \texttt{mongodb\_play.py}.

\subsection{Concrete Examples}\label{sec:hastopics}

We used the implementation to replay a complex mobile manipulation (kitting) task \cite{han2020towards}, including picking, navigation, and placement. As shown in Figure \ref{fig:pick}, a Fetch robot successfully replayed its past arm movement and manipulation of a misrecognized object. An AR visualization, the green projection, was also replayed to indicate the grasped object in the past. In Figure \ref{fig:nav}, the robot replayed its past navigation behavior around an obstacle, with yellow projection for the obstacle and arrows for the detour path. As another manipulation example (Figure \ref{fig:place}), Fetch replayed its object placement arm movement with AR projection onto the section of caddy that was misrecognized.

Here, we list the topics we have identified and recorded, with the Fetch robot in mind. Although different robots use different topics, it is particularly easy to find corresponding topics. For body (arm and wheel) movement, the following topics were replayed:
\begin{itemize}[noitemsep,topsep=0pt]
\item Gripper: ``/gripper\_controller/gripper\_action/goal''
\item Arm and torso: ``/arm\_with\_torso\_controller/follow\_joint\\\_trajectory/goal''
\item Torso: ``/torso\_controller/follow\_joint\_trajectory/goal''
\item Head: ``/head\_controller/point\_head/goal''
\item Head: ``/head\_controller/follow\_joint\_trajectory/goal''
\item Wheel movement: ``/cmd\_vel''
\end{itemize}

Indeed, any topics can be replayed. For example, the green projection in Figure \ref{fig:pick} and the yellow projection (Figure \ref{fig:nav}) to indicate ground obstacle is a PointCloud2 message in the sensor\_msgs package\footnote{\url{https://wiki.ros.org/sensor_msgs}}. The arrow in Figure \ref{fig:nav} is a Path message from the nav\_msgs package\footnote{\url{https://wiki.ros.org/nav_msgs}}. The white projection for the caddy section is a Marker message provided by the rviz package\footnote{\url{https://wiki.ros.org/rviz/DisplayTypes/Marker}}.
In addition to visuals, non-visual messages can also be replayed. For example, we have been able to replay speech by replaying messages from ``/robotsound'', used by the sound\_play package\footnote{\url{https://wiki.ros.org/sound_play}}.

\section{Evaluation}\label{sec:eval}

We have also validated the effectiveness and efficiency of robot behavior replay for past behavior explanation using our implementation in the scenarios described in Figure \ref{fig:pick}--\ref{fig:place}. In an experiment (N=665) we reported in detail in another paper \cite{han2022past}, we assessed a combination of different replays: Physical replay, AR projection replay, and speech replay. The combination of these three modalities has achieved the best overall effectiveness, helping participants infer where the robot grasped the misrecognized object (Manipulation inference; Figure \ref{fig:pick}), why the robot made a detour (Navigation inference; Figure \ref{fig:nav}), and which section of the caddy it wrongly placed an object into (Placement inference; Figure \ref{fig:place}).
For efficiency, 60\% of the participants were able to infer the detour reason with AR projection replay only. Speech replay alone helped participants to quickly make inference on the caddy section. For manipulation inference, although more participants can quickly make the inference with projection or speech reply, more participants also reported they were not able to get the answer from such replays. While physical replay is time-consuming, it ensured inference accuracy.
Trust and workload were also rated by participants, but there were no conclusive findings. More details about this experiment can be found in \cite{han2022past}.

\section{Limitations}

Although we were able to implement both manipulation and navigation replays, the navigation replay has approximately a few centimeter error when reaching the destination position in the case of Figure \ref{fig:nav}. Although this may seem short, it could be problematic where a following dependent behavior, e.g., the placement replay shown in Figure \ref{fig:place}, will not have the desired state, e.g., the gripper may not be above the correct caddy section, as seen in the fourth photo in Figure \ref{fig:place}. In such cases, we may solve this issue by recording intermediate goals as how body movement was recorded, shown in Section \ref{sec:hastopics}.

\section{Conclusion}

In this work, we presented an implementation of robot behavior replay. We justified the choices of database for robotic data storage and robot task representation for specifying and separating robot behaviors. Figure \ref{fig:workflow} shows the required components and steps to replay robot behaviors. We also discussed the wide applicability of this technique, i.e., capability to replay any kind of ROS messages.

\section*{Acknowledgments}

This work has been supported in part by the Office of Naval Research (N00014-18-1-2503) and the National Science
Foundation (IIS-1909864). We also thank Vittoria Santoro and Jenna Parrillo for code help.

\bibliography{bib}

\begin{thebibliography}{37}
\providecommand{\natexlab}[1]{#1}

\bibitem[{Amir, Doshi-Velez, and Sarne(2018)}]{amir2018agent}
Amir, O.; Doshi-Velez, F.; and Sarne, D. 2018.
\newblock Agent strategy summarization.
\newblock In \emph{Proceedings of the 17th International Conference on
  Autonomous Agents and MultiAgent Systems}, 1203--1207.

\bibitem[{Beetz et~al.(2018)Beetz, Be{\ss}ler, Haidu, Pomarlan,
  Bozcuo{\u{g}}lu, and Bartels}]{beetz2018know}
Beetz, M.; Be{\ss}ler, D.; Haidu, A.; Pomarlan, M.; Bozcuo{\u{g}}lu, A.~K.; and
  Bartels, G. 2018.
\newblock Know rob 2.0—a 2nd generation knowledge processing framework for
  cognition-enabled robotic agents.
\newblock In \emph{2018 IEEE International Conference on Robotics and
  Automation (ICRA)}, 512--519. IEEE.

\bibitem[{Beetz, M{\"o}senlechner, and Tenorth(2010)}]{beetz2010cram}
Beetz, M.; M{\"o}senlechner, L.; and Tenorth, M. 2010.
\newblock CRAM—A Cognitive Robot Abstract Machine for everyday manipulation
  in human environments.
\newblock In \emph{2010 IEEE/RSJ International Conference on Intelligent Robots
  and Systems}, 1012--1017. IEEE.

\bibitem[{Beetz, Tenorth, and Winkler(2015)}]{beetz2015open}
Beetz, M.; Tenorth, M.; and Winkler, J. 2015.
\newblock {Open-EASE}--a knowledge processing service for robots and
  robotics/{AI} researchers.
\newblock In \emph{2015 IEEE International Conference on Robotics and
  Automation (ICRA)}, 1983--1990. IEEE.

\bibitem[{Bohren and Cousins(2010)}]{bohren2010smach}
Bohren, J.; and Cousins, S. 2010.
\newblock The smach high-level executive.
\newblock \emph{IEEE Robotics \& Automation Magazine}, 17(4): 18--20.

\bibitem[{Brunner et~al.(2016)Brunner, Steinmetz, Belder, and
  D{\"o}mel}]{brunner2016rafcon}
Brunner, S.~G.; Steinmetz, F.; Belder, R.; and D{\"o}mel, A. 2016.
\newblock RAFCON: A graphical tool for engineering complex, robotic tasks.
\newblock In \emph{2016 IEEE/RSJ International Conference on Intelligent Robots
  and Systems (IROS)}, 3283--3290. IEEE.

\bibitem[{Colledanchise and {\"O}gren(2018)}]{colledanchise2018behavior}
Colledanchise, M.; and {\"O}gren, P. 2018.
\newblock \emph{Behavior trees in robotics and AI: An introduction}.
\newblock CRC Press.

\bibitem[{French et~al.(2019)French, Wu, Pan, Zhou, and
  Jenkins}]{french2019learning}
French, K.; Wu, S.; Pan, T.; Zhou, Z.; and Jenkins, O.~C. 2019.
\newblock Learning behavior trees from demonstration.
\newblock In \emph{2019 International Conference on Robotics and Automation
  (ICRA)}, 7791--7797. IEEE.

\bibitem[{Han et~al.(2020{\natexlab{a}})Han, Allspaw, LeMasurier, Parrillo,
  Giger, Ahmadzadeh, and Yanco}]{han2020towards}
Han, Z.; Allspaw, J.; LeMasurier, G.; Parrillo, J.; Giger, D.; Ahmadzadeh,
  S.~R.; and Yanco, H.~A. 2020{\natexlab{a}}.
\newblock Towards mobile multi-task manipulation in a confined and integrated
  environment with irregular objects.
\newblock In \emph{2020 IEEE International Conference on Robotics and
  Automation (ICRA)}, 11025--11031. IEEE.

\bibitem[{Han et~al.(2021)Han, Giger, Allspaw, Lee, Admoni, and
  Yanco}]{han2021building}
Han, Z.; Giger, D.; Allspaw, J.; Lee, M.~S.; Admoni, H.; and Yanco, H.~A. 2021.
\newblock Building the foundation of robot explanation generation using
  behavior trees.
\newblock \emph{ACM Transactions on Human-Robot Interaction (THRI)}, 10(3):
  1--31.

\bibitem[{Han et~al.(2022)Han, Parrillo, Wilkinson, Yanco, and
  Williams}]{han2022projecting}
Han, Z.; Parrillo, J.; Wilkinson, A.; Yanco, H.~A.; and Williams, T. 2022.
\newblock Projecting Robot Navigation Paths: Hardware and Software for
  Projected AR.
\newblock In \emph{Proceedings of the 2022 ACM/IEEE International Conference on
  Human-Robot Interaction}, 623--628.

\bibitem[{Han, Phillips, and Yanco(2021)}]{han2021need}
Han, Z.; Phillips, E.; and Yanco, H.~A. 2021.
\newblock The need for verbal robot explanations and how people would like a
  robot to explain itself.
\newblock \emph{ACM Transactions on Human-Robot Interaction (THRI)}, 10(4):
  1--42.

\bibitem[{Han, Rygina, and Williams(2022)}]{han2022evaluating}
Han, Z.; Rygina, P.; and Williams, T. 2022.
\newblock Evaluating Referring Form Selection Models in Partially-Known
  Environments.
\newblock In \emph{Proceedings of the 15th International Natural Language
  Generation Conference}.

\bibitem[{Han et~al.(2020{\natexlab{b}})Han, Wilkinson, Parrillo, Allspaw, and
  Yanco}]{han2020projection}
Han, Z.; Wilkinson, A.; Parrillo, J.; Allspaw, J.; and Yanco, H.~A.
  2020{\natexlab{b}}.
\newblock Projection mapping implementation: Enabling direct externalization of
  perception results and action intent to improve robot explainability.
\newblock In \emph{Proceedings of the AI-HRI Symposium at AAAI-FSS 2020}.

\bibitem[{Han and Yanco(Under review)}]{han2022past}
Han, Z.; and Yanco, H.~A. Under review.
\newblock Communicating Missing Causal Information to Explain a Robot's Past
  Behavior.
\newblock Under review.

\bibitem[{Hayes and Shah(2017)}]{hayes2017improving}
Hayes, B.; and Shah, J.~A. 2017.
\newblock Improving robot controller transparency through autonomous policy
  explanation.
\newblock In \emph{2017 12th ACM/IEEE International Conference on Human-Robot
  Interaction (HRI}, 303--312. IEEE.

\bibitem[{Ikeda and Szafir(2022)}]{ikeda2022advancing}
Ikeda, B.; and Szafir, D. 2022.
\newblock Advancing the Design of Visual Debugging Tools for Roboticists.
\newblock In \emph{Proceedings of the 2022 ACM/IEEE International Conference on
  Human-Robot Interaction}, 195--204.

\bibitem[{Lim, Baumgarten, and Colton(2010)}]{lim2010evolving}
Lim, C.-U.; Baumgarten, R.; and Colton, S. 2010.
\newblock Evolving behaviour trees for the commercial game DEFCON.
\newblock In \emph{European conference on the applications of evolutionary
  computation}, 100--110. Springer.

\bibitem[{Macenski et~al.(2020)Macenski, Mart{\'\i}n, White, and
  Clavero}]{macenski2020marathon}
Macenski, S.; Mart{\'\i}n, F.; White, R.; and Clavero, J.~G. 2020.
\newblock The marathon 2: A navigation system.
\newblock In \emph{2020 IEEE/RSJ International Conference on Intelligent Robots
  and Systems (IROS)}, 2718--2725. IEEE.

\bibitem[{Nakawala et~al.(2018)Nakawala, Goncalves, Fiorini, Ferringo, and
  De~Momi}]{nakawala2018approaches}
Nakawala, H.; Goncalves, P.~J.; Fiorini, P.; Ferringo, G.; and De~Momi, E.
  2018.
\newblock Approaches for action sequence representation in robotics: A review.
\newblock In \emph{2018 IEEE/RSJ International Conference on Intelligent Robots
  and Systems (IROS)}, 5666--5671. IEEE.

\bibitem[{Niemueller, Lakemeyer, and Srinivasa(2012)}]{niemueller2012generic}
Niemueller, T.; Lakemeyer, G.; and Srinivasa, S.~S. 2012.
\newblock A generic robot database and its application in fault analysis and
  performance evaluation.
\newblock In \emph{2012 IEEE/RSJ International Conference on Intelligent Robots
  and Systems}, 364--369. IEEE.

\bibitem[{OpenRobotics(2022)}]{rosshowcase}
OpenRobotics. 2022.
\newblock {Official ROS robot showcase}.
\newblock \url{https://robots.ros.org}.
\newblock Accessed: 2022-07-13.

\bibitem[{Palamara et~al.(2008)Palamara, Ziparo, Iocchi, Nardi, Lima, and
  Costelha}]{palamara2008robotic}
Palamara, P.; Ziparo, V.; Iocchi, L.; Nardi, D.; Lima, P.; and Costelha, H.
  2008.
\newblock A robotic soccer passing task using petri net plans (demo paper).
\newblock In \emph{Proc. of 7th Int. Conf. on Autonomous Agents and Multiagent
  Systems (AAMAS 2008)}, 1711--1712.

\bibitem[{Paxton et~al.(2017)Paxton, Hundt, Jonathan, Guerin, and
  Hager}]{paxton2017costar}
Paxton, C.; Hundt, A.; Jonathan, F.; Guerin, K.; and Hager, G.~D. 2017.
\newblock CoSTAR: Instructing collaborative robots with behavior trees and
  vision.
\newblock In \emph{2017 IEEE international conference on robotics and
  automation (ICRA)}, 564--571. IEEE.

\bibitem[{Quigley et~al.(2009)Quigley, Conley, Gerkey, Faust, Foote, Leibs,
  Wheeler, Ng et~al.}]{quigley2009ros}
Quigley, M.; Conley, K.; Gerkey, B.; Faust, J.; Foote, T.; Leibs, J.; Wheeler,
  R.; Ng, A.~Y.; et~al. 2009.
\newblock ROS: an open-source Robot Operating System.
\newblock In \emph{ICRA workshop on open source software}, volume~3, 5.

\bibitem[{RethinkRobotics(2022)}]{intera}
RethinkRobotics. 2022.
\newblock {Intera Software Platform for Industrial Automation}.
\newblock \url{https://www.rethinkrobotics.com/intera}.
\newblock Accessed: 2022-07-26.

\bibitem[{Rosen et~al.(2019)Rosen, Whitney, Phillips, Chien, Tompkin,
  Konidaris, and Tellex}]{rosen2019communicating}
Rosen, E.; Whitney, D.; Phillips, E.; Chien, G.; Tompkin, J.; Konidaris, G.;
  and Tellex, S. 2019.
\newblock Communicating and controlling robot arm motion intent through
  mixed-reality head-mounted displays.
\newblock \emph{The International Journal of Robotics Research}, 38(12-13):
  1513--1526.

\bibitem[{Sagredo-Olivenza et~al.(2017)Sagredo-Olivenza, G{\'o}mez-Mart{\'\i}n,
  G{\'o}mez-Mart{\'\i}n, and Gonz{\'a}lez-Calero}]{sagredo2017trained}
Sagredo-Olivenza, I.; G{\'o}mez-Mart{\'\i}n, P.~P.; G{\'o}mez-Mart{\'\i}n,
  M.~A.; and Gonz{\'a}lez-Calero, P.~A. 2017.
\newblock Trained behavior trees: Programming by demonstration to support ai
  game designers.
\newblock \emph{IEEE Transactions on Games}, 11(1): 5--14.

\bibitem[{Scheutz et~al.(2019)Scheutz, Williams, Krause, Oosterveld, Sarathy,
  and Frasca}]{scheutz2019overview}
Scheutz, M.; Williams, T.; Krause, E.; Oosterveld, B.; Sarathy, V.; and Frasca,
  T. 2019.
\newblock An overview of the distributed integrated cognition affect and
  reflection diarc architecture.
\newblock \emph{Cognitive architectures}, 165--193.

\bibitem[{Schneider(1990)}]{schneider1990implementing}
Schneider, F.~B. 1990.
\newblock Implementing fault-tolerant services using the state machine
  approach: A tutorial.
\newblock \emph{ACM Computing Surveys (CSUR)}, 22(4): 299--319.

\bibitem[{Stange et~al.(2022)Stange, Hassan, Schr{\"o}der, Konkol, and
  Kopp}]{stange2022self}
Stange, S.; Hassan, T.; Schr{\"o}der, F.; Konkol, J.; and Kopp, S. 2022.
\newblock Self-Explaining Social Robots: An Explainable Behavior Generation
  Architecture for Human-Robot Interaction.
\newblock \emph{Frontiers in Artificial Intelligence}, 87.

\bibitem[{Stange and Kopp(2020)}]{stange2020effects}
Stange, S.; and Kopp, S. 2020.
\newblock Effects of a social robot’s self-explanations on how humans
  understand and evaluate its behavior.
\newblock In \emph{2020 15th ACM/IEEE International Conference on Human-Robot
  Interaction (HRI)}, 619--627. IEEE.

\bibitem[{Tenorth and Beetz(2009)}]{tenorth2009knowrob}
Tenorth, M.; and Beetz, M. 2009.
\newblock KnowRob—knowledge processing for autonomous personal robots.
\newblock In \emph{2009 IEEE/RSJ international conference on intelligent robots
  and systems}, 4261--4266. IEEE.

\bibitem[{Walker et~al.(2022)Walker, Phung, Chakraborti, Williams, and
  Szafir}]{walker2022virtual}
Walker, M.; Phung, T.; Chakraborti, T.; Williams, T.; and Szafir, D. 2022.
\newblock Virtual, augmented, and mixed reality for human-robot interaction: A
  survey and virtual design element taxonomy.
\newblock \emph{arXiv preprint arXiv:2202.11249}.

\bibitem[{Wise et~al.(2016)Wise, Ferguson, King, Diehr, and
  Dymesich}]{wise2016fetch}
Wise, M.; Ferguson, M.; King, D.; Diehr, E.; and Dymesich, D. 2016.
\newblock Fetch and freight: Standard platforms for service robot applications.
\newblock In \emph{Workshop on autonomous mobile service robots}.

\bibitem[{Zhu and Williams(2020)}]{zhu2020effects}
Zhu, L.; and Williams, T. 2020.
\newblock Effects of proactive explanations by robots on human-robot trust.
\newblock In \emph{International Conference on Social Robotics}, 85--95.
  Springer.

\bibitem[{Ziparo et~al.(2008)Ziparo, Iocchi, Nardi, Palamara, and
  Costelha}]{ziparo2008petri}
Ziparo, V.~A.; Iocchi, L.; Nardi, D.; Palamara, P.~F.; and Costelha, H. 2008.
\newblock Petri net plans: a formal model for representation and execution of
  multi-robot plans.
\newblock In \emph{Proceedings of the 7th international joint conference on
  Autonomous agents and multiagent systems-Volume 1}, 79--86.

\end{thebibliography}

\end{document}